\documentclass[journal]{IEEEtran}
\usepackage[normalem]{ulem}
\usepackage[pagebackref=true,breaklinks=true,colorlinks,bookmarks=false]{hyperref}
\usepackage[pdftex]{graphicx}
\usepackage{array}
\usepackage{xcolor}
\usepackage{multirow}
\usepackage{cite}
\usepackage{amsmath}
\usepackage[caption=false,font=footnotesize]{subfig}
\usepackage{url}
\usepackage{amsfonts}
\usepackage{pifont}

\usepackage{adjustbox}

\newlength\fsdurthree

\begin{document}

\title{Zero shot framework for satellite image restoration}
\author{Praveen Kandula, and~A. N. Rajagopalan, \textit{Senior Member, IEEE}}

\markboth{Journal of \LaTeX\ Class Files,~Vol.~14, No.~8, August~2015}%
{Shell \MakeLowercase{\textit{et al.}}: Bare Demo of IEEEtran.cls for IEEE Journals}

\maketitle
\begin{abstract}
Satellite images are typically subject to multiple distortions. Different factors affect the quality of satellite images, including changes in atmosphere, surface reflectance, sun illumination, viewing geometries etc., limiting its application to downstream tasks. In supervised networks, the availability of paired datasets is a strong assumption. Consequently, many unsupervised algorithms have been proposed to address this problem. These methods synthetically generate a large dataset of degraded images using image formation models. A neural network is then trained with an adversarial loss to discriminate between images from distorted and clean domains. However, these methods yield suboptimal performance when tested on real images that do not necessarily conform to the generation mechanism. Also, they require a large amount of training data and are rendered unsuitable when only a few images are available. We propose a distortion disentanglement and knowledge distillation framework for satellite image restoration to address these important issues. Our algorithm requires only two images: the distorted satellite image to be restored and a reference image with similar semantics. Specifically, we first propose a mechanism to disentangle distortion. This enables us to generate images with varying degrees of distortion using the disentangled distortion and the reference image. We then propose the use of knowledge distillation to train a restoration network using the generated image pairs. As a final step, the distorted image is passed through the restoration network to get the final output. Ablation studies show that our proposed mechanism successfully disentangles distortion. Exhaustive experiments on different timestamps of Google-Earth images and publicly available datasets, LEVIR-CD and SZTAKI, show that our proposed mechanism can tackle a variety of distortions and outperforms existing state-of-the-art restoration methods visually as well as on quantitative metrics.
\end{abstract}
\IEEEpeerreviewmaketitle


\section{Introduction and Related works}
\label{sec:intro}

Satellite imagery is used for different applications, including extracting mineral deposits, agriculture development, disaster mitigation, weather forecast etc. Unfortunately, the quality of the captured images is affected by intrinsic properties of acquisition systems, and imaging conditions. While the intrinsic properties include physical condition of sensors, detectors, camera tilts etc., the quality of satellite images is also affected by terrain, atmospheric bending, sun illumination, and viewing angles during the time of capture. The combined effect of these distortions (or degradations) is difficult to model mathematically, rendering the existing satellite restoration works \cite{sat_res1, satres3, bi2021haze, ref1, ref2, ref3, minor1, minor2, minor3} unsuitable for real-world images. 
We use the words `distortions' and `degradations' interchangeably in this paper.

Learning-based deep networks give impressive results for individual restoration tasks like deblurring \cite{vasu2018non, mohan2021deep, nimisha2018unsupervised, rao2014harnessing, rajagopalan1998optimal, paramanand2011depth, purohit2020region, vasu2017local, nimisha2018generating, suin2020spatially, purohit2019bringing, mohan2019unconstrained, purohit2019efficient, suin2021gated, purohit2022multi, varghese2022fast}, dehazing \cite{ancuti2019ntire, purohit2019multilevel, dehaze2, mandal2019local}, inpainting \cite{inpaint2, suin2021distillation}, enhancement \cite{kandula2023illumination, el2020aim}, super-resolution \cite{suin2020degradation, rajagopalan2003motion, purohit2021spatially, suresh2007robust, bhavsar2010resolution, purohit2020mixed, nimisha2021blind}, bokeh rendering \cite{purohit2019depth}, inpainting \cite{suin2021distillation, bhavsar2012range} etc. Unsupervised methods relax the need for paired datasets for training a deep net. In this setting, a large dataset of clean and distorted images is required without the need to enforce the criterion of paired images. These methods typically have CycleGAN \cite{CycleGAN} as a backbone and discriminate images from clean and distorted domains using cycle consistency and adversarial losses. This formulation has been successfully employed for different restoration tasks like denoising \cite{unsuperviseddenoise, rajagopalan2005background}, super-resolution \cite{CycleinCycle}, shadow-removal \cite{unsupervisedshadow}, etc., with suitable modifications in loss functions and network architecture. Although paired images are not required, pooling a large number of distorted images remains a challenge. To tackle this, unsupervised methods resort to synthesizing distortions using different formulations akin to supervised techniques. 


In this paper, we propose a learning-based algorithm to restore a distorted satellite image using a reference image. We propose a distortion disentanglement and knowledge distillation mechanism for image restoration, which addresses the shortcomings of previous works. Although our method requires a reference image, this is not difficult to acquire in satellite imaging, where images are taken at regular intervals. The captured image might have undesirable artifacts due to low light, weather conditions etc., but there is a high probability that at least one past image is distortion-free. Using reference image as input, recent supervised super-resolution networks \cite{refSR1,refSR2} showed improved performance. Our algorithm can be considered as an extension of existing zero-shot and reference-based algorithms. Different from zero-shot methods, our method uses a reference image in addition to an input image during training.
\\The main contributions of our work are summarized below:

\begin{itemize}
    \item To the best of our knowledge, this is the first-ever learning-based approach to explicitly separate distortion information from a degraded satellite image using only a pair of images (distorted and reference images).  
    \item We propose feature adversarial loss and feature regularization loss to train distortion disentanglement network (DDN). Different ablation studies show that our design architecture and loss functions in DDN effectively decouple content and distortion from an input image.
    \item A unique strength of our method is that it is not distortion-specific i.e., the proposed algorithm restores satellite images with unknown and challenging distortions. Also, our algorithm does not warrant the reference and distorted images to be exactly aligned.
\end{itemize}

\begin{figure*}[t!]
	\centering
	\includegraphics[width=0.95\linewidth]{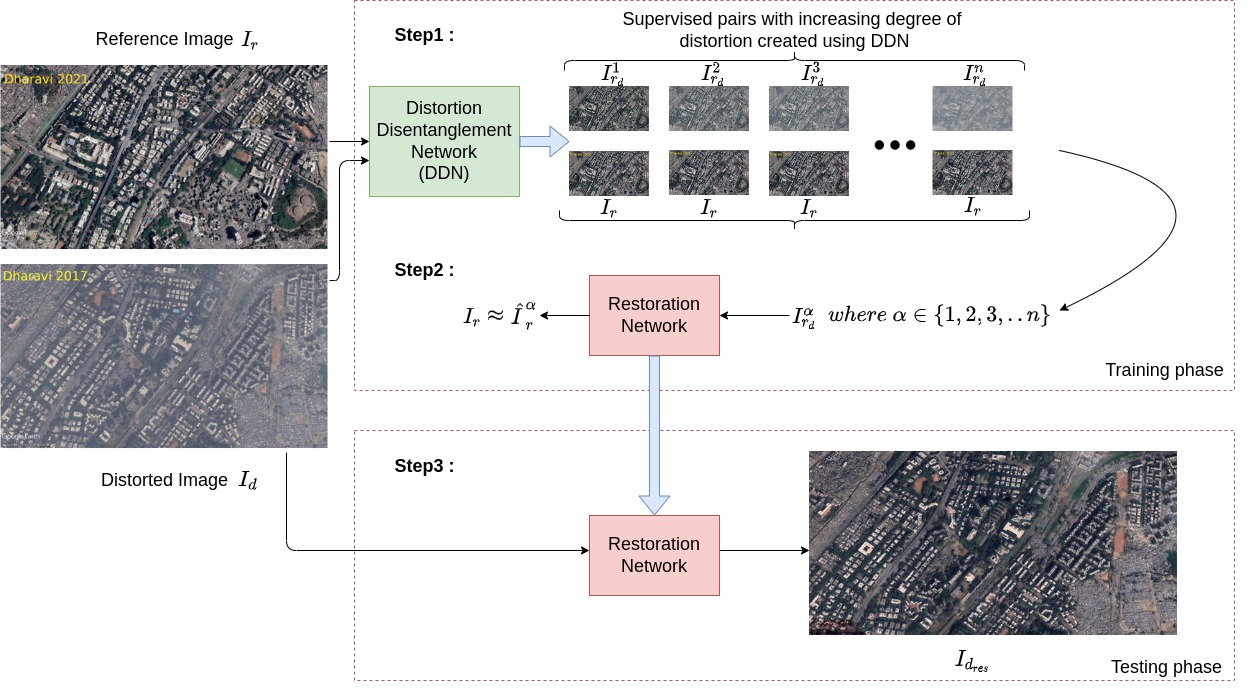}
	\caption{Proposed network architecture. Given a reference image ($I_r$) and distorted image ($I_d$), the objective is to restore $I_d$ using $I_r$. Here, $I_r$ and $I_d$ are real satellite images from 2017 and 2021, respectively, from Dharavi (Mumbai). The proposed restoration mechanism involves three steps. In step1, a distortion disentanglement network (DDN) is used to decouple the content and distortion information from $I_d$. Supervised pairs with increasing degrees of distortion are created using decoupled information. In the second step, a restoration network is trained using the supervised pairs. As the last step, the distorted image ($I_d$) is passed through the trained restoration network for the final restored image  ($I_{d_{res}}$). A major highlight of our algorithm is that the proposed model needs only a pair of images ($I_r$ and $I_d$) for restoration. Also, a unique strength of our algorithm is that it is not distortion specific i.e., since the DDN couples distortion from input images, our model can handle challenging and unknown distortions.}
	\label{Network_architecture}
\end{figure*}



\section{Proposed Method}
\label{Sec: Proposed_method}
The objective of our method is to restore a distorted satellite image $I_d$ using a clean reference image $I_r$. For conciseness, the distorted satellite image is referred to as distorted image. Also, the content in $I_r$ is not required to perfectly match that of $I_d$, i.e., $I_r$ and $I_d$ can be captured at different time stamps. 

\begin{figure*}[t]
\setlength{\tabcolsep}{1pt}
\scriptsize
\centering
\begin{tabular}{ccccccccccccc}

 \includegraphics[width=.125\linewidth]{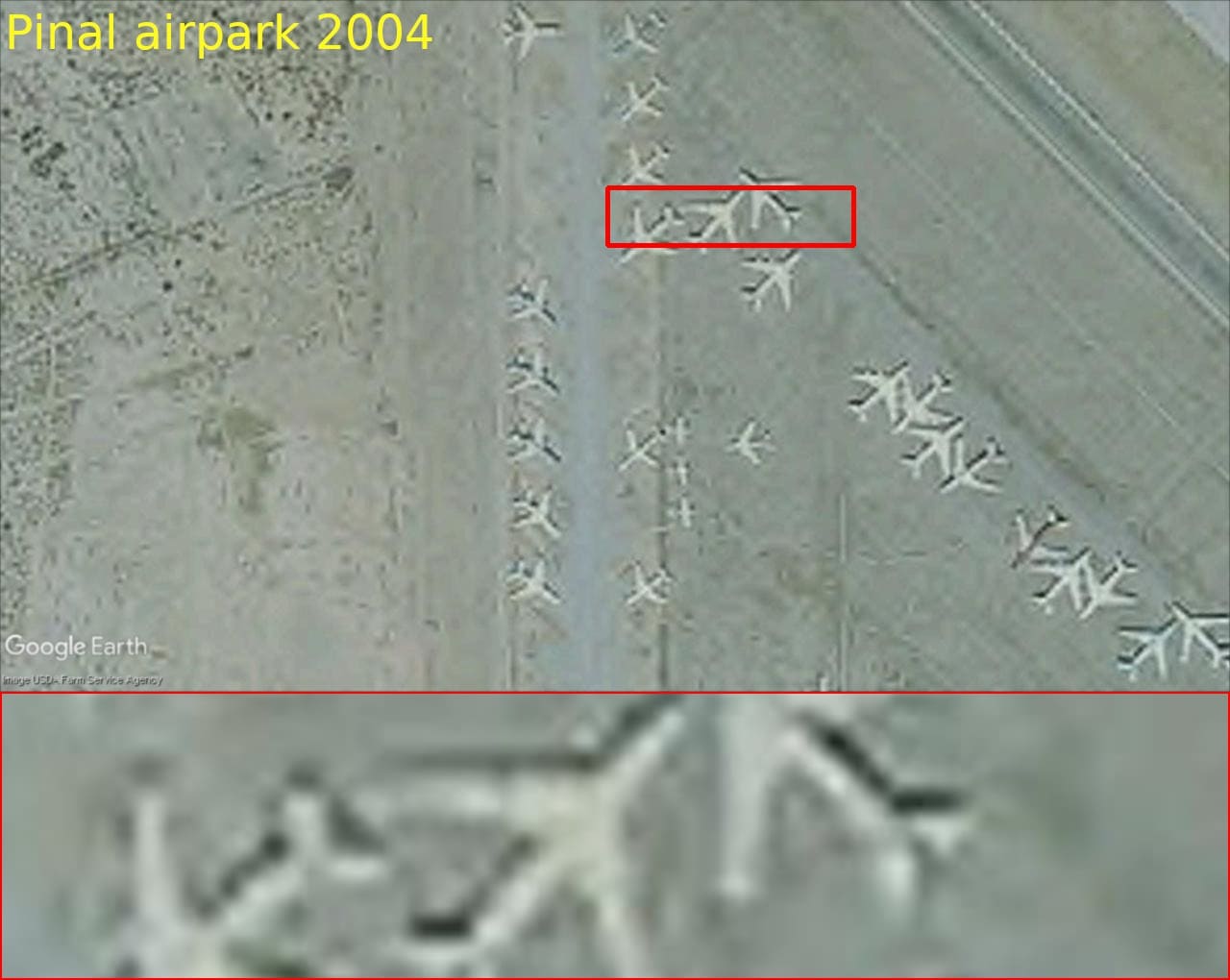} &
 \includegraphics[width=.125\linewidth]{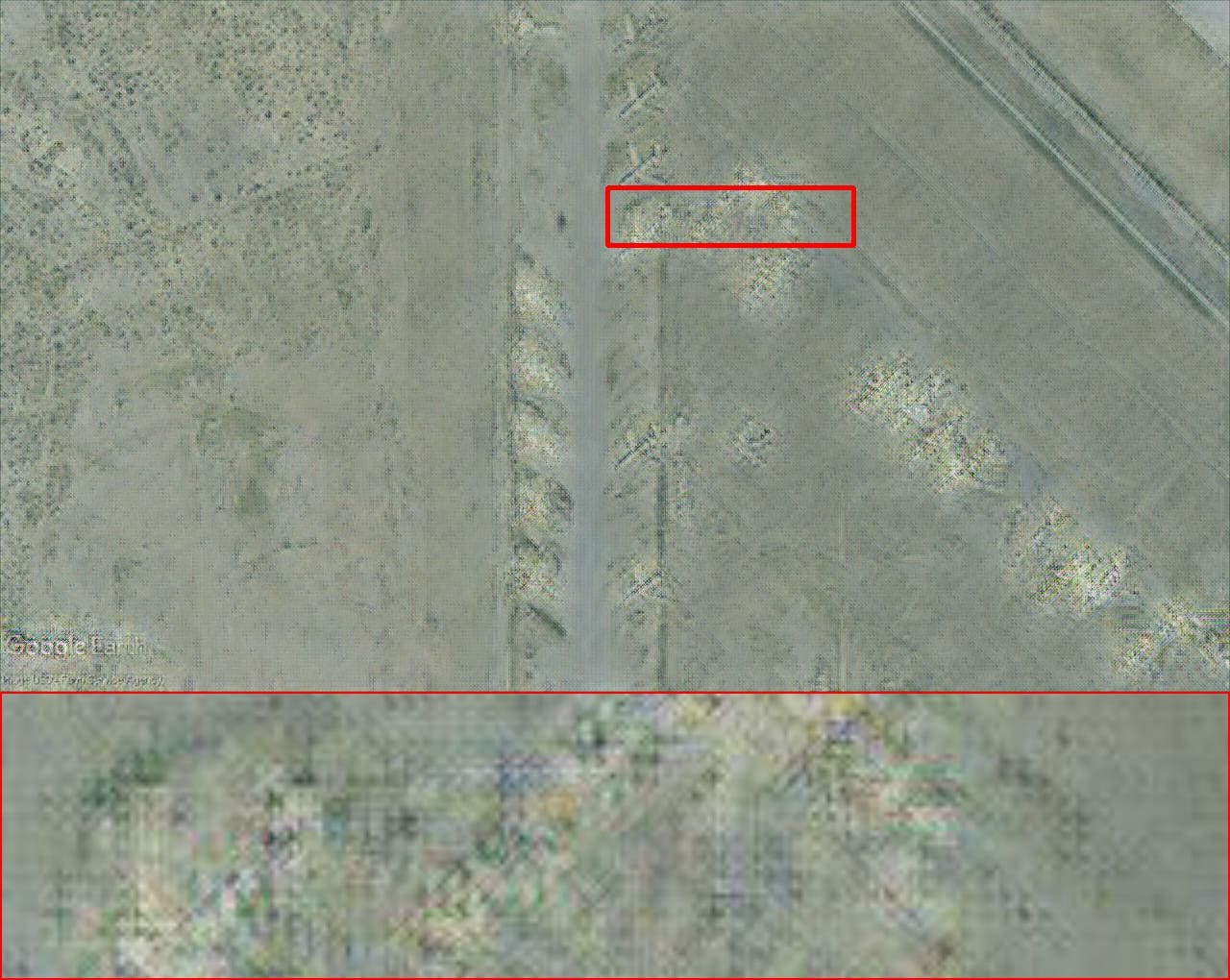} &
 \includegraphics[width=.125\linewidth]{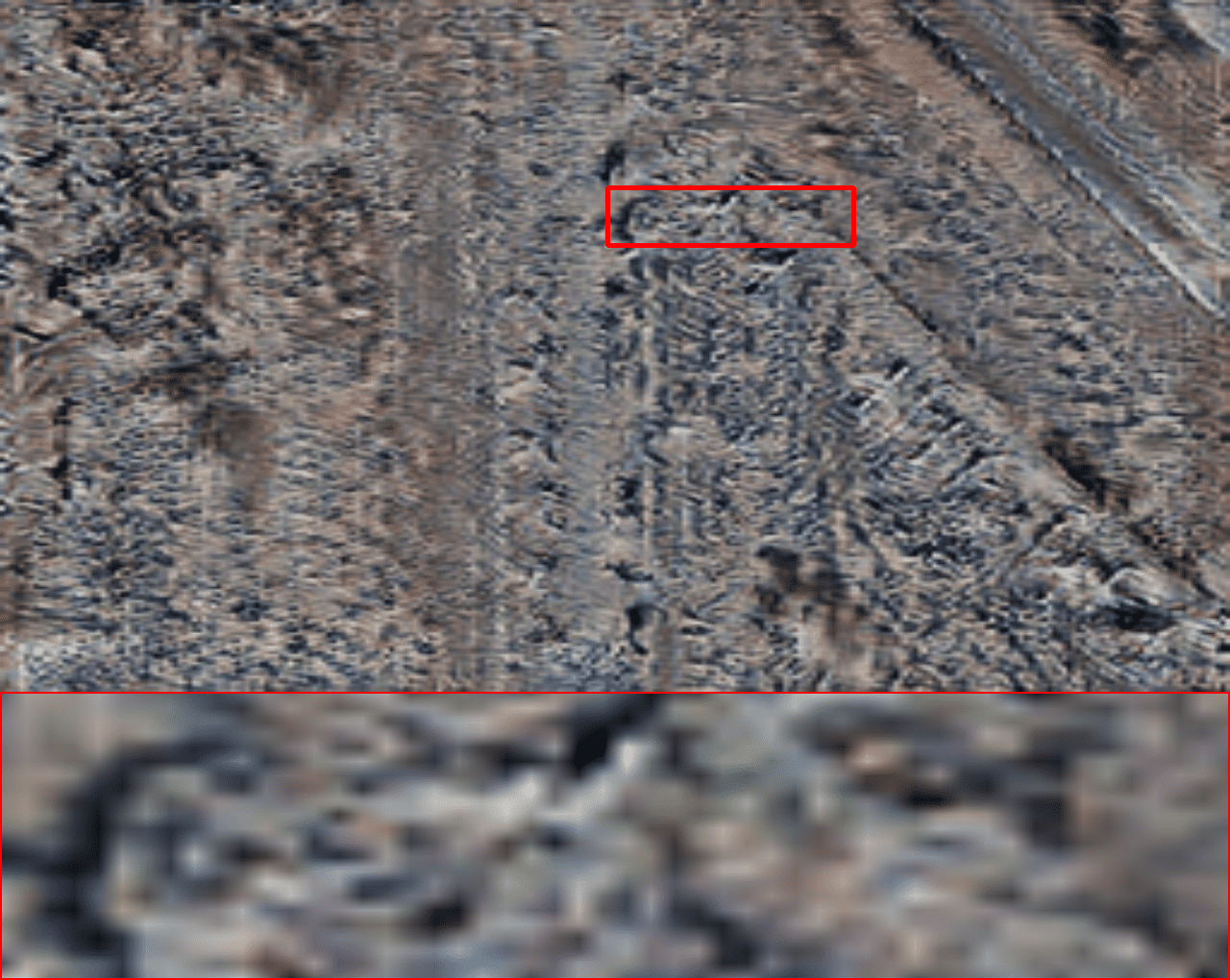} &
 \includegraphics[width=.125\linewidth]{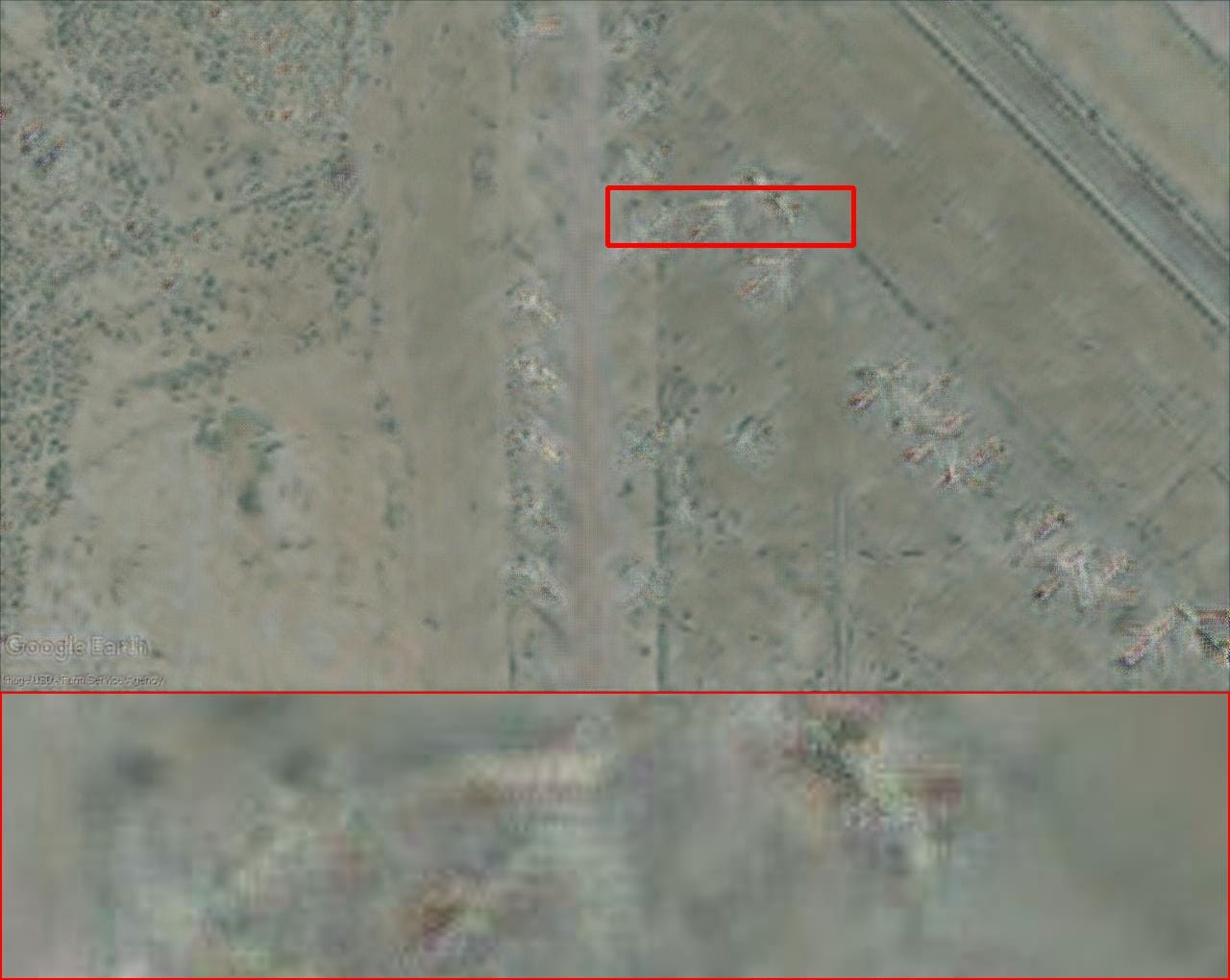} &
 
 \includegraphics[width=.125\linewidth]{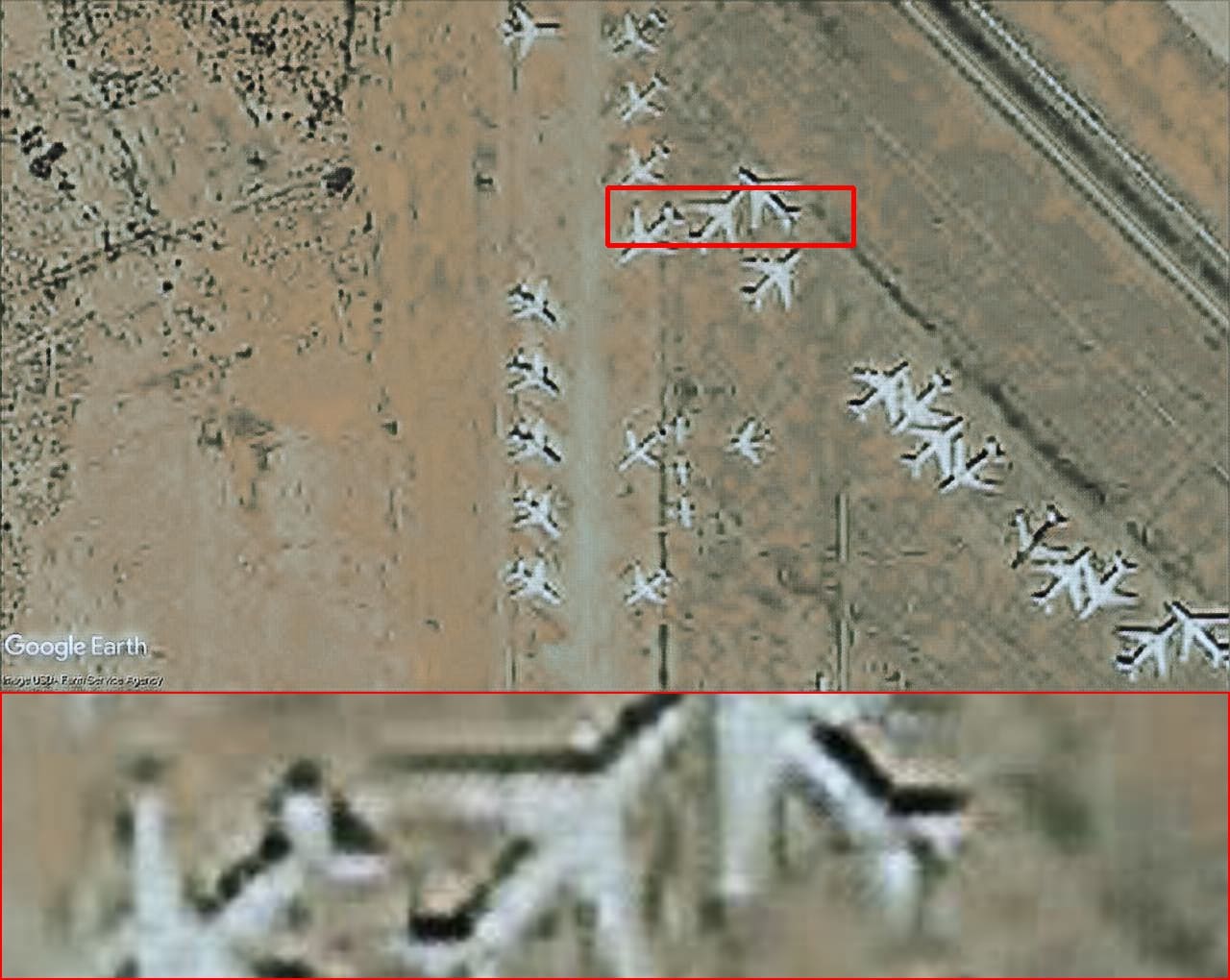} &
 \includegraphics[width=.125\linewidth]{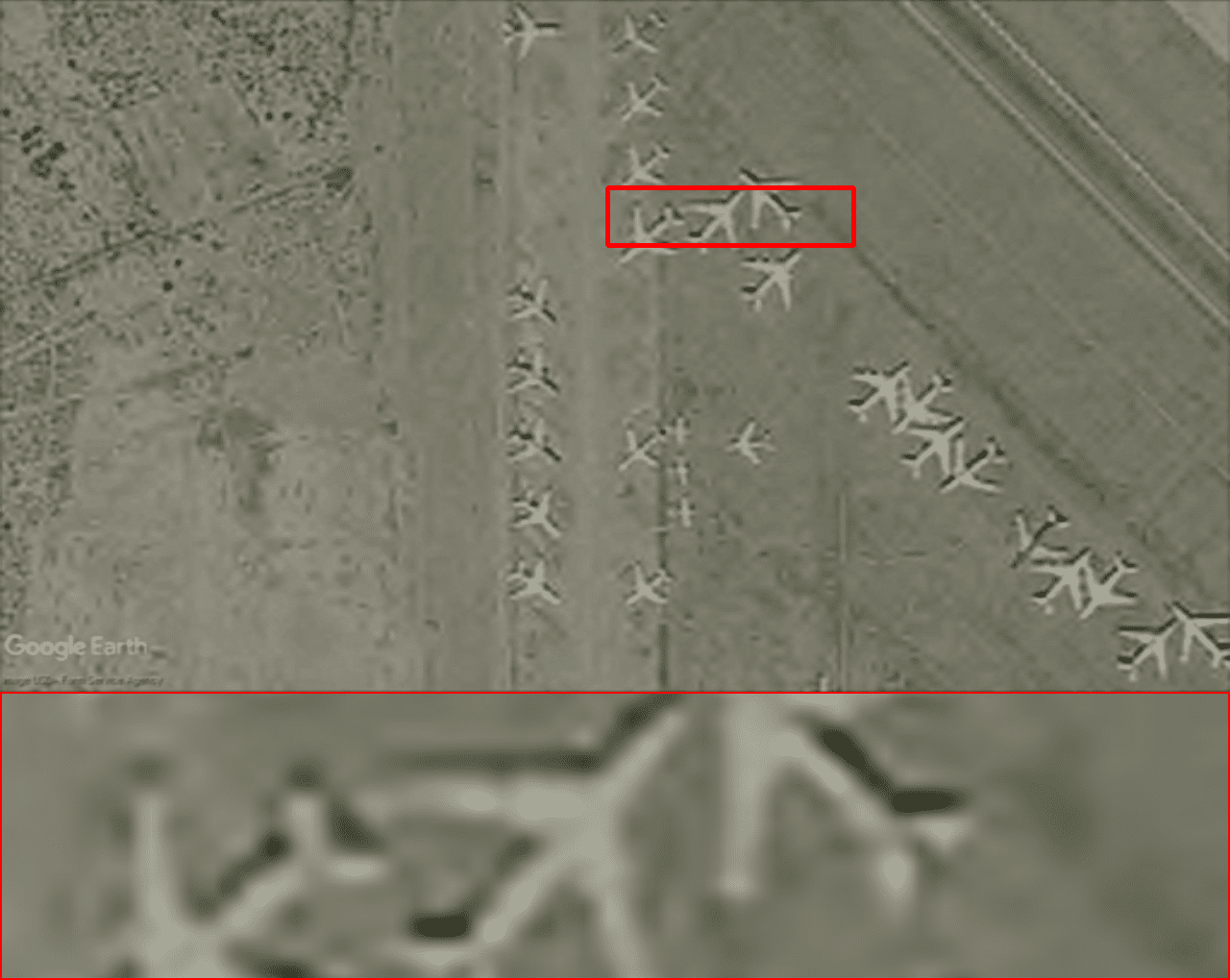} &
 \includegraphics[width=.125\linewidth]{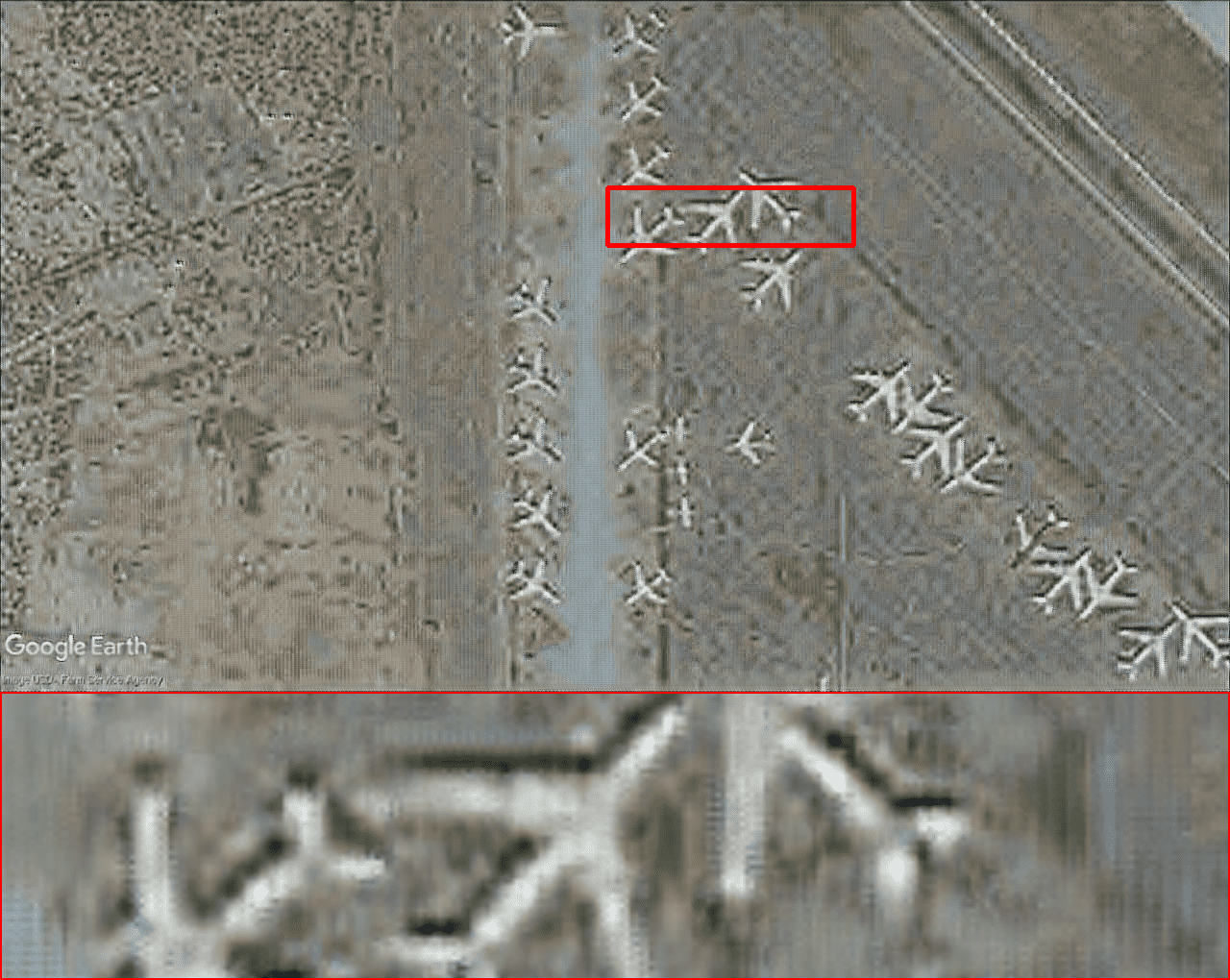}&
 \includegraphics[width=.125\linewidth]{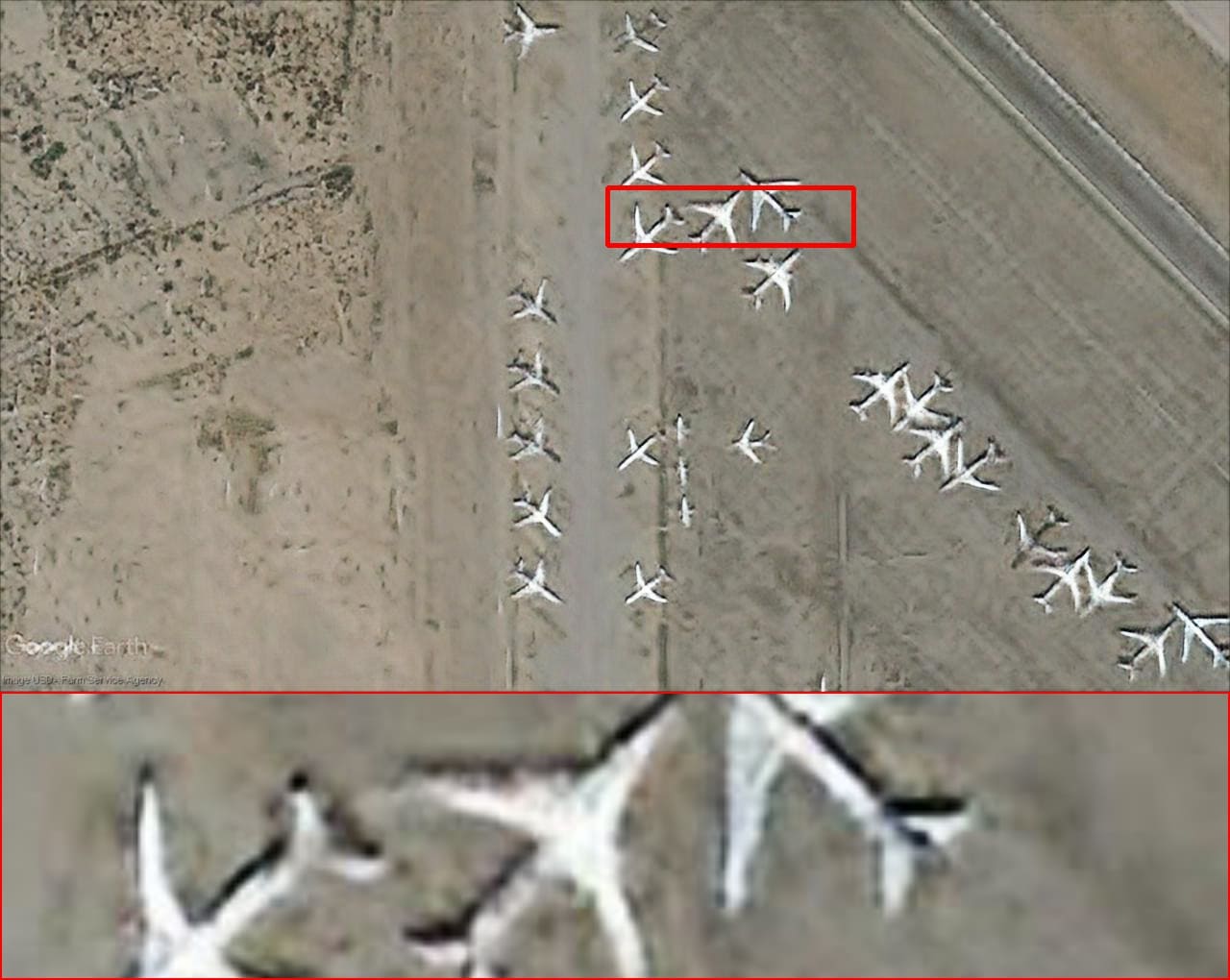}\\
 
  Distorted image &  VGGSR\cite{Supple_SR} &  DCLGAN\cite{Supple_cyclegan} &  DHMP \cite{super_comp} & 
  UDSD \cite{chellapa} & UEGAN \cite{Supple_enhance} &  CycleGAN \cite{CycleGAN} & Ours \\
   
 

\end{tabular}\\ 
\caption{Visual comparisons with state-of-the-art restoration methods on real satellite images. The distorted image in the top row (1.(a)) suffers from blur artifacts and has green colour tingle all over the image. Similarly, distorted image in third row (2. (a)) also has colour cast and contents are not clearly visible. Competing methods clearly failed to restore the missing details and remove the colour tinge. Even in the CycleGAN (1.(g)), green colour is still present and airplane structure is deformed. On the other hand, results of our method completely removed the unwanted colours and the contents are clearly visible.  }
\label{fig: Comparisons}
\end{figure*}

Since these images share similar scene semantics, one possible approach is to use supervised networks with $I_r$ and $I_d$ as paired images over pixel-wise loss functions. However, the restored image is noisy and lacks structural coherence (see Fig. \ref{fig: Comparisons}, third row). This can be attributed to two reasons. First, although $I_r$ and $I_d$ share scene semantics, they are not necessarily aligned. For instance, the aircrafts in $I_r$ is geometrically off by a significant margin compared to $I_d$. These misalignments can badly affect the outcome for supervised loss functions. In addition, since $I_d$ and $I_r$ are imaged at different timestamps, there can be significant differences in their image contents. Nonetheless, supervised loss functions fit the non-existent information in the distorted image to the one present in reference images resulting in bad test results.

 \subsection{Disentanglement of distortion}
 \label{Sec: Disenatlgment of distortion}
 Given a reference image ($I_r$) and distorted image ($I_d$), the objective is to disentangle distortion information from $I_d$ using $I_r$. This is a challenging task as we are provided with only two images. DDN has three sub networks, a content encoder $E_c$, a distortion encoder $E_d$, and a decoder $D$. Given the distorted image $I_d$, $E_c$ extracts only content information, while $E_d$ extracts only distortion information. Since $I_r$ does not contain any distortion information, $E_d$ does not extract any information from it. Given $I_d$(or $I_r$), $D$ uses the combined feature maps of $E_c$  and $E_d$ to estimate the corresponding image.  To ensure that $E_c$ extracts only content information from the given image, we adopt the following strategy.
 
\begin{itemize}
    \item Since $I_r$ is free from distortion, $E_c$ must ideally extract only content information. Let the output of $E_c$ given $I_r$ be denoted as $F_{r_{c}} \in R ^{1\times C \times H \times W}$, where $C, H, W$ are number of channels, height and width of feature map $F_{r_{c}}$, respectively.
    \item Given $I_d$, $E_c$ should extract only content features. Let the output of $E_c$ given $I_d$ be denoted as  $F_{d_{c}} \in R ^{1\times C \times H \times W}$, where $C, H, W$ are as mentioned above. In order to ensure that $E_c$ extracts only content information, we propose to use adversarial loss on feature maps. The feature maps $F_{r_{c}}$ are used as real samples while $F_{d_{c}}$ are the fake samples. The loss function for this objective is termed feature adversarial loss and is formulated as 
\begin{equation}
\begin{split}
 \mathbb{L}_{adv}(E_c, E_d, D, D_{adv}) =  \mathbb{E}[\log (D_{adv}(F_{r_{c}}))] + \\
 \mathbb{E} [\log(1 - D_{adv}(F_{d_{c}}))]  
\end{split}
\end{equation}
\end{itemize}

\begin{figure*}[t]
\setlength{\tabcolsep}{1pt}
\scriptsize
\centering
\begin{tabular}{ccccccccccc}

\includegraphics[width=0.45\linewidth]{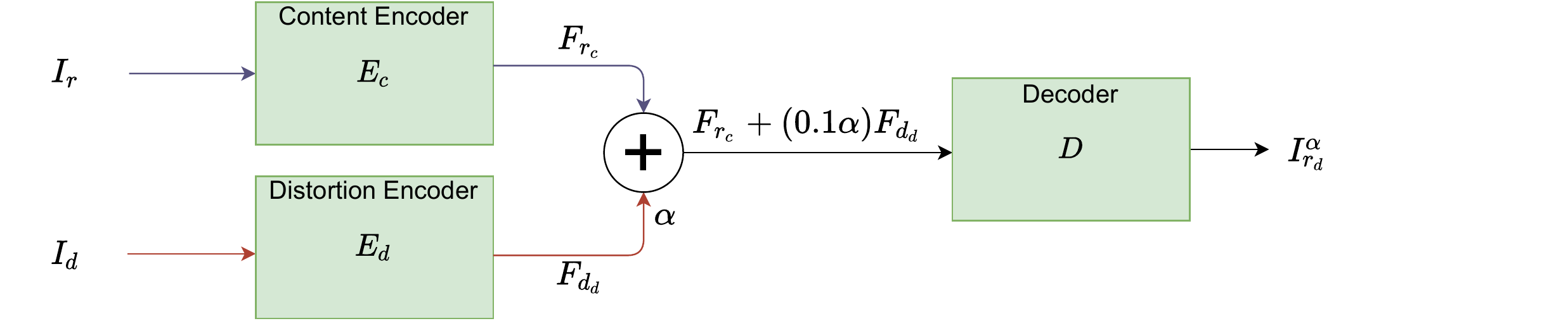} &
 \includegraphics[width=0.45\linewidth]{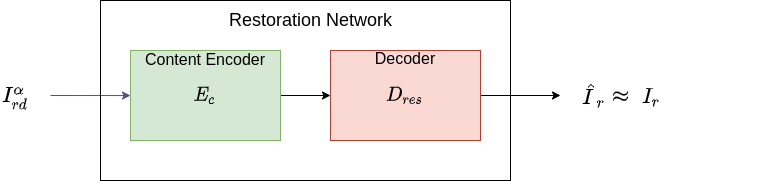}
\\
(a) Overview of distortion transfer mechanism. & (b) Restoration using knowledge distillation. \\
\end{tabular}\\ 

\caption{ (a) Given $I_r$ and $I_d$, the objective is to transfer distortion from $I_d$ to $I_r$. To achieve this, feature maps of $E_c$ given $I_r$ ($F_{r_{c}}$) is combined with feature maps of $E_d$ given $I_d$ ($F_{d_{d}}$) and passed through the decoder network. A detailed explanation of the distortion transfer mechanism is given in Sec. \ref{Sec: distortion transfer}. (b) Instead of using a vanilla encoder-decoder network for training restoration network, $E_c$ from distortion disentanglement network (DNN) is used as encoder network, which improves the convergence rate.  }
\label{Distortion_trasnfer}
\end{figure*}

\subsection{Feature regularization loss}
The motivation for this loss function is to ensure that $E_d$ does not respond to $I_r$, since $I_r$ is distortionless. We achieve this by constraining all the output feature maps of $E_d$ to zero. This also ensures that $E_d$ extracts only distortion information from $I_d$. This loss function is formulated as 
\begin{equation}
 L_{reg} = ||\sum_{i=1}^{\rho} F^{i}_{r_{d}}||_1
\end{equation}
where $F^{i}_{r_{d}}$ is the output feature map of $E_d$ with $I_r$ as input and $i$ denotes intermediate feature maps of $E_d$.
\subsection{Cyclic loss functions}
In order to ensure that the Decoder $D$ is effectively trained to recover the input image, we use cyclic loss functions on $I_r$ and $I_d$. Given $I_d$, $E_c$ encodes only content information while $E_d$ encodes only distortion due to $L_{adv}$ and $L_{reg}$. $D$ takes combined feature maps of $E_c$ and $E_d$ to get back the input image $I_d$. The cyclic loss function for $I_d$ can be defined as 
\begin{equation}
 L_{d_{cy}} = ||D(E_c(I_d) + E_d(I_d)) - I_d||_1
\end{equation}
Similarly, cyclic loss for $I_r$ is defined as 
\begin{equation}
 L_{r_{cy}} = ||D(E_c(I_r) + E_d(I_r)) - I_r||_1
\end{equation}
where $L_{d_{cy}}$ and $L_{r_{cy}}$ refer to distortion cyclic loss and reference cyclic loss functions, respectively. While $L_{adv}$ and $L_{reg}$ ensure that $E_c$ extracts only content information, the cyclic losses $L_{d_{cy}}$ and $L_{r_{cy}}$ force $E_d$ to learn distortion information from input images in addition to estimating the corresponding input image. 

The total loss function used in our disentanglement framework is the combination of the above loss functions. All the loss functions are used within a single iteration, and the four networks ($E_c$, $E_d$, $D$, $D_{adv}$) are trained end-to-end. The final loss function for our distortion disentanglement network methodology can be written as 
\begin{equation}
\label{Eq: total_loss}
 L_{Total} = \lambda_1 L_{adv} + \lambda_2 L_{reg} + \lambda_3 L_{d_{cy}} + \lambda_4 L_{r_{cy}}
\end{equation}
where $\lambda _1$, $\lambda _2$, $\lambda _3$, and $\lambda _4$ refer to trade-off weights balancing different loss functions.

\subsection{Distortion transfer}
\label{Sec: distortion transfer}
Once the DDN is trained using $L_{Total}$, the modules $E_{c}$ and $E_{d}$ extract only content and distortion information, respectively, from any input image. To transfer the distortion from $I_{d}$ to $I_r$, we use the distortion transfer mechanism illustrated in Fig. \ref{Distortion_trasnfer} (a) and formulated as 
\begin{equation}
\label{Eq: distortion_transfer}
 I_{r_{d}}^{\alpha} = D(F_{r_{c}} + (0.1\alpha) F_{d_{d}})
\end{equation}
where $F_{r_{c}}$ represents the output feature maps of $E_c$ given $I_r$,  $F_{d_{d}}$ the output feature maps of $E_d$ for $I_d$ image,  $I_{r_{d}}^\alpha$ is the result of the distortion transfer mechanism and $\alpha \in 1,2...n $. The objective of $\alpha$ is to create varying degrees of distortion on $I_r$. Using this mechanism, a dataset of supervised pairs, $(I_{r_{d}}^{1}, I_r), (I_{r_{d}}^{2}, I_r) ... (I_{r_{d}}^{n}, I_r)$ is created. The key idea behind using different values of $\alpha$ is to train a robust restoration network that can effectively restore $I_d$.

\subsection{Restoration by knowledge distillation}
We propose to use the supervised pairs (created by distortion transfer) within a knowledge distillation (KD) framework to restore $I_d$ effectively.
Inspired by acceleration techniques \cite{KD} (a type of KD) that transfer the convolutional filters from one model to another, we take advantage of the information learned during the disentanglement process. Specifically, we use an encoder-decoder model for restoration, with encoder being the pretrained network $E_c$ (from DDN) that can judiciously extract content features from input images. We fix the encoder model and update the parameters of the decoder ($D_{res}$) alone using MSE loss as shown in Fig. \ref{Distortion_trasnfer} (b). This can be formulated as

\begin{equation}
 \hat{I}^{\alpha}_{r} = D_{res}(E_c(I_{r_{d}}^{\alpha}))
\end{equation}
where $\hat{I}^{\alpha}_{r}$ is the output of the restoration network given $I_{r_{d}}^{\alpha}$. The main advantage of using a KD mechanism is that the restoration network can be accelerated and efficiently trained. The convergence time using the proposed mechanism is significantly reduced compared to a network that trains from scratch. 

Note that DDN disentangles distortion in $I_d$, while the restoration network is efficiently trained to tackle that distortion. The restoration network is trained robustly by varying $\alpha$ in Eq. 6 to effectively handle different amounts of distortion. As a final step, $I_d$ is passed through the trained restoration network to obtain the output restored image, as shown in Fig. \ref{Network_architecture}. This can be written as 
\begin{equation}
 I_{d_{res}} = D_{res}(E_c(I_d))
\end{equation}
where $I_{d_{res}}$ is the final restored output given $I_d$ as input to the restoration network.
\section{Experiments}
The experiments section is arranged as follows: (i) implementation details, (ii) dataset and competing methods, (iii)  ablation studies, (iv) metrics used, quantitative comparisons, and (v) visual comparisons.

\subsection{Implementation details}
We use Pytorch library to train both DDN and restoration networks. We first train DDN using Eq. \ref{Eq: total_loss} with weights for different loss functions set as $\lambda _1$= 1, $\lambda _2$= 10, $\lambda _3$= 1 and $\lambda _4$= 1. Then a supervised dataset with varying amounts of distortion is created from the distortion transfer mechanism using Eq. \ref{Eq: distortion_transfer} with $n$ = 100, i.e., hundred distorted images are created to train the restoration network. The DDN network is trained for 4000 iterations, while the restoration network is trained for 150 epochs. For both DDN and restoration network, we use the following options: ADAM optimizer to update the model parameters, momentum values with $\beta _1$ = 0.9 and $\beta _2$ = 0.99, patch size of 512 (original size of all the images is 1280x780) and a learning rate of 0.0001. Inference and convergence times for DDN and restoration network are discussed in supplementary. VGGSR \cite{Supple_SR} showed that the features extracted from initial layers of VGG16 \cite{VGG}  can be used for different downstream tasks such as denoising and super-resolution.

We mainly compare our method with unsupervised methods CycleGAN\cite{CycleGAN}, DCLGAN \cite{Supple_cyclegan},  UDSD \cite{chellapa}, UEGAN \cite{Supple_enhance} and two supervised restoration methods DHMP \cite{super_comp}, and VGGSR \cite{Supple_SR}. Additionally, we evaluated the performance of our model on two publicly available datasets, LEVIR-CD \cite{ge2}, and SZTAKI \cite{ge3}. 

\section{Conclusion}

We proposed a deep learning framework for restoring a distorted satellite image using a reference image. We showed that content and distortion present in the image can be effectively decoupled. This allowed us to transfer the distortion to a clean image and create distortion-clean image pairs. With these supervised pairs, a restoration network was efficiently trained using knowledge distillation. Extensive ablation studies were conducted to reveal the effect of the loss functions used in our network. The performance of our method is superior to prior art both visually and on standard image quality assessment metrics. The restoration results reveal that our algorithm can successfully handle images suffering from multiple distortions.



{\small
	\bibliographystyle{IEEEtran}
	\bibliography{egbib}
}
\end{document}